\documentclass[10pt,twocolumn,letterpaper]{article}

\usepackage{iccv}
\usepackage{times}
\usepackage{epsfig}
\usepackage{graphicx}
\usepackage{subfigure}
\usepackage{amsmath}
\usepackage{amssymb}
\usepackage{mathrsfs}
\usepackage{multirow}
\usepackage{booktabs}
\usepackage{makecell}
\usepackage{boldline}
\usepackage{xcolor}
\usepackage{authblk}
\usepackage[resetlabels,labeled]{multibib}
\newcites{Ref}{References}
\usepackage{amsmath}
\usepackage{amsfonts}
\usepackage{amssymb}
\usepackage{amsthm}
\usepackage{bm}
\usepackage{bbm}
\usepackage{mathtools}
\usepackage{enumitem}
\usepackage{thmtools,thm-restate}
\usepackage[ruled,vlined]{algorithm2e}
\SetAlFnt{\small}
\usepackage{color}
\usepackage{graphicx}
\usepackage{comment}
\theoremstyle{plain}

\theoremstyle{definition}

\theoremstyle{remark}

\def\BB{\mathcal{B}}\def\CC{\mathcal{C}}
\def\DD{\mathcal{D}}
\def\II{\mathcal{I}}
\def\LL{\mathcal{L}}
\def\NN{\mathcal{N}}

\def\VV{\mathcal{V}}\def\WW{\mathcal{W}}\def\XX{\mathcal{X}}

\def\Rbb{\mathbb{R}}

\def\R{\Rbb}

\usepackage{etex,etoolbox}
\usepackage{environ}

\makeatletter
\providecommand{\@fourthoffour}[4]{#4}
\newcommand\fixstatement[2][\proofname\space of]{%
	\ifcsname thmt@original@#2\endcsname
	\AtEndEnvironment{#2}{%
		\xdef\pat@label{\expandafter\expandafter\expandafter
			\@fourthoffour\csname thmt@original@#2\endcsname\space\@currentlabel}%
		\xdef\pat@proofof{\@nameuse{pat@proofof@#2}}%
	}%
	\else
	\AtEndEnvironment{#2}{%
		\xdef\pat@label{\expandafter\expandafter\expandafter
			\@fourthoffour\csname #1\endcsname\space\@currentlabel}%
		\xdef\pat@proofof{\@nameuse{pat@proofof@#2}}%
	}%
	\fi
	\@namedef{pat@proofof@#2}{#1}%
}

\newcounter{proofcount}

\NewEnviron{proofatend}{%
	\edef\next{%
		\noexpand\begin{proof}[\pat@proofof\space\pat@label]%
			\unexpanded\expandafter{\BODY}}%
		\global\toks\numexpr\prooftoks+\value{proofcount}\relax=\expandafter{\next\end{proof}}
	\stepcounter{proofcount}}

\def\printproofs{%
	\count@=\z@
	\loop
	\the\toks\numexpr\prooftoks+\count@\relax
	\ifnum\count@<\value{proofcount}%
	\advance\count@\@ne
	\repeat}
\makeatother

\fixstatement{lemma}
\fixstatement{theorem}
\fixstatement{proposition}
\fixstatement{corollary}

\usepackage[pagebackref=true,breaklinks=true,letterpaper=true,colorlinks,bookmarks=false]{hyperref}

\iccvfinalcopy % *** Uncomment this line for the final submission

\def\iccvPaperID{3345} % *** Enter the ICCV Paper ID here

\ificcvfinal\fi

\makeatletter
\renewcommand\AB@affilsepx{, \protect\Affilfont}
\makeatother

\newcommand{\SKIP}[1]{}

\begin{document}

%%%%%%%%% TITLE
%%%%%%%%% TITLE
\title{Learning to Find Common Objects Across Few Image Collections}
\author[1]{Amirreza Shaban\thanks{Equal contribution. Contact at \texttt{amirreza@gatech.edu} or \texttt{amir.rahimi@anu.edu.au}.}}
\author[2]{Amir Rahimi$^{*}$} %\footnotemark[1] did not work
\author[1]{Shray Bansal}
\author[2]{\\Stephen Gould}
\author[1]{Byron Boots}
\author[2]{Richard Hartley}

\affil[1]{Georgia Tech}
\affil[2]{ACRV, ANU Canberra}

\maketitle
\ificcvfinal\thispagestyle{empty}\fi

%%%%%%%%% ABSTRACT
\begin{abstract}
%We address the problem of finding images of a common object across a collection of image proposals. 
Given a collection of bags where each bag is a set of images, our goal is to select one image from each bag such that the selected images are from the same object class.
We model the selection as an energy minimization problem with unary and pairwise potential functions. Inspired by recent few-shot learning algorithms, we propose an approach to learn the potential functions directly from the data. Furthermore, we propose a fast greedy inference algorithm for energy minimization. We evaluate our approach on few-shot common object recognition as well as object co-localization tasks. Our experiments show that learning the pairwise and unary terms greatly improves the performance of the model over several well-known methods for these tasks. The proposed greedy optimization algorithm achieves performance comparable to state-of-the-art structured inference algorithms while being $\sim$10 times faster.

\end{abstract}
\vspace{-.3cm}
%%%%%%%%% BODY TEXT
\begin{figure}[t]
    \centering
    \includegraphics[width=1.00\columnwidth]{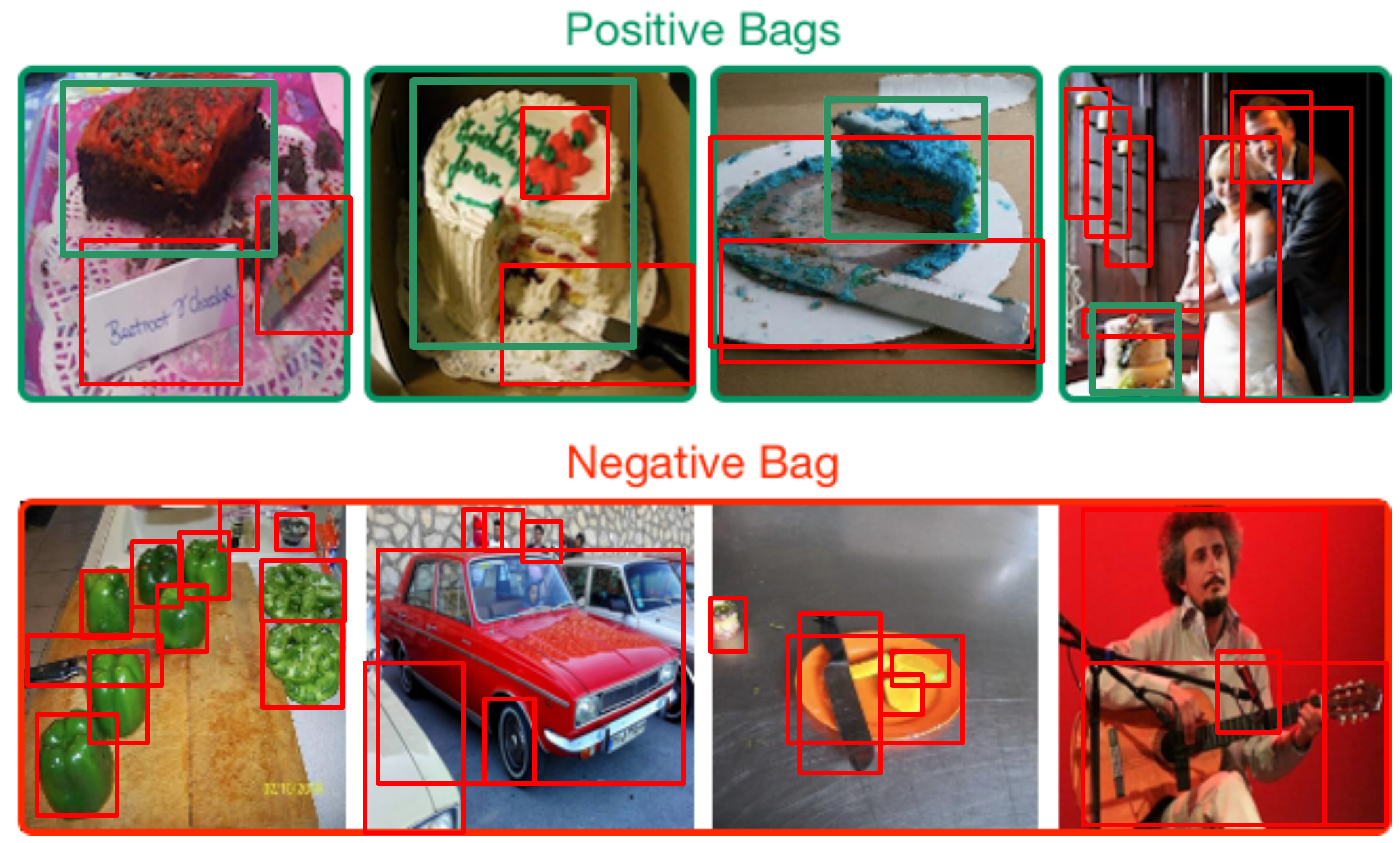}
    \caption{\label{fig:summary}\em Co-localization, shown here, is an instance of the general problem of finding common objects addressed in this paper. Each image in the top row generates a positive bag containing a set of cropped regions from that image. The task is to find a common object from the positive bags by selecting one region from each image (green bounding boxes). Cropped regions from the images in the bottom row form a negative bag as they do not contain the common object. The negative bag is optional here but can reduce ambiguity. For example, since a knife is present in the negative bag it can not be the desired common object.}
    \vspace{-1em}
\end{figure}
\section{Introduction}
We address the problem of finding images of a common object across bags of images. -
%We address the problem of finding images containing objects from a common but unknown class, given a collection of image proposals. 
The input is a collection of bags, each containing several images from multiple classes. 
A bag is labelled as \emph{positive} with respect to a given object class if it contains at least one image from that class and \emph{negative} 
if none of the images in the bag are from the object class. 
The task is to find an instance of the common object in each positive bag. It is not assumed that objects of the common class have been seen previously during training.
% shray: this is a little bit confusing probably because the problem is difficult to explain. I don't think you make any mistakes in explaining it. But I think I get confused a little because here you talk about a bag having multiple images and then a paragraph later bags are proposals. 

Since collections of images may accidentally contain irrelevant common objects (for instance indoor images often contain person), the purpose of a negative bag is to indicate objects we are {\em not} looking for, but which may be common to the positive
bags.

%\paragraph{Motivation. } 
Several computer vision problems, including co-segmentation, co-localization, and unsupervised video object tracking and segmentation  %~\cite{fu2014object,babenko2009visual} 
have been formulated in this way~\cite{vicente2011object,faktor2013co,Hsu_2018_ECCV, fu2014object,babenko2009visual}. 
In the co-localization problem, Figure \ref{fig:summary}, each bag contains many cropped image regions (object proposals) from one image.
%and these bags are labelled as positive or negative based on the presence of the common object. 
The goal is to identify proposals, one per positive bag, that contain the common object. We design our approach to address the general problem of finding common objects from positive bags and evaluate it on two problems: few-shot common object recognition and object co-localization.

Weakly supervised classification methods like multiple-instance learning~\cite{maron1998framework} have been used to address this type of problems, but they require many training bags to learn new concepts~\cite{pmlr-v80-ilse18a}. Meta-learning techniques~\cite{Finn2017ModelAgnosticMF, santoro2016meta, shaban2018truncated} have been shown to reduce the need for training instances in few-shot learning, but these methods require full supervision for the new classes.

%MRF
We model the problem of finding common objects as 
a minimum-energy graph labelling problem,
otherwise known as a bidirectional graphical model or Markov Random Field.
Each node of the graph
corresponds to a positive bag and %labels correspond to the choice of one image from each bag. 
a graph labelling corresponds to choosing one image in each positive bag, the goal being to find a labelling that contains the common object. 
We use the word {\em selection}
instead of {\em labelling} to refer to the process of selecting
one image from each bag.
The energy minimization problem uses unary and pairwise potential functions, where unary potentials reflect the relation of images in the positive bags to the images in the negative bag and the pairwise potentials derive from a 
similarity measure between pairs of images from two positive bags.
%Pairwise terms are defined for all pairs of nodes (bags),
%so the graphical model is defined on the complete graph.
The unary and pairwise potentials are computed using similar,
but separately trained networks. We adapt the relation network~\cite{sung2018learning}, which has been successfully used in few-shot recognition to compute pairwise potentials, and
propose a new algorithm that uses the relationship of an image to all of the images in the negative bag
to provide unary potentials.

Once unary and pairwise potentials have been computed,
a simple merge-and-prune inference heuristic is used to 
find a minimum-cost labelling. This provides a simple but
effective solution to the NP-hard problem of optimal
graph labelling. %The proposed greedy search algorithm for finding the common object is based on the following observation: \textit{the common object for the complete problem is a common object in any subset of the bags as well.}

Although graphical models have been used for Multiple Instance Learning~(MIL) problems~\cite{Deselaers2010,Hajimirsadeghi:2013:MIL:3023638.3023665}, our method %generalizes to novel classes by learning the potential functions by using a meta-learning approach. % shray: made a change check if it's correct
uses a learning-based approach, inspired by meta-learning, to increase the generalization power of potential functions to novel classes.

%Our contribution in this paper has two parts: 
%\paragraph{Contributions. } 
We make the following contributions:
\begin{enumerate}
\item We introduce a method to transfer knowledge from large-scale strongly supervised datasets by learning pairwise and unary potentials and demonstrate the superiority of this learned relation metric to earlier MIL approaches on two problems.
 
 \item We propose a specialized %fast greedy 
 algorithm for structured inference that achieves % and show that this method achieves
 comparable performance to state-of-the-art inference methods while requiring less computational time.
 \end{enumerate}

%We make the following two contributions. (1) Introduce a learning methodology to provide pairwise and unary terms from large-scale strongly supervised datasets. We validate the superiority of the learning approach to previously proposed MIL approaches that do not use a learnable relation metric. (2) Propose a specialized fast greedy algorithm for structured inference. We show that this method can achieve comparable performance to the state-of-the-art inference methods while requiring less computational time.
\section{Related Work} \label{sec:review}
Multiple instance learning~(MIL)~\cite{carbonneau2018multiple,pathakICLR15} methods have been used for learning weakly supervised tasks such as object localization (WSOL)~\cite{Jie2017DeepSL,chen2017discover,zhuang2017attend,shen2018generative}. In a standard MIL framework, instance labels in each positive bag are treated as hidden variables with the constraint that at least one of them should be positive. MI-SVM and mi-SVM~\cite{andrews2003support} are two popular methods for MIL, and have been widely adapted for many weakly supervised computer vision problems, achieving state-of-the-art results in many different applications~\cite{carbonneau2018multiple, Doran:2014:TEA:2666867.2666935}. In these methods, images in each bag inherit the label of the bag and an SVM is trained to classify images. The trained SVM is used to relabel the instances and this process is repeated until the labels remain stable. While in MI-SVM only the image with the highest score in positive bags are labeled as positive, mi-SVM allows more than one positive label in each positive bag in the relabeling process.

Co-saliency~\cite{zhang2015co,Hsu_2018_ECCV}, co-segmentation~\cite{vicente2011object,faktor2013co,hochbaum2009efficient}, and co-localization~\cite{li2016image} methods have the same kind of output as WSOL methods. Similar to standard MIL algorithms, some of these methods rely on a relatively large training set for learning novel classes~\cite{li2016image, tang2014co}. The main difference between these methods and WSOL methods is that they usually do not utilize negative examples~\cite{vicente2011object, li2016image, tang2014co}. Negative examples in our method are optional and could be used to improve the results of the co-localization task.

Our approach is related to weakly supervised methods that make use of auxiliary fully-labelled data to accelerate the learning of new categories~\cite{46626, Hoffman2015DetectorDI,Shi2017WeaklySO,rochan2015weakly,deselaers2012weakly}. Since visual classes share many visual characteristics, knowledge from fully-labelled source classes is used to learn from the weakly-labelled target classes. 
The general approach is to use the labelled dataset to learn an embedding function for images and use MI-SVM to classify instances of the weakly labelled dataset in this space~\cite{46626, Hoffman2015DetectorDI, Shi2017WeaklySO}. We show that learning a scoring function to compare images in the embedded space significantly improves the performance of this approach, especially when few positive images are available.
Rochan et al.~\cite{rochan2015weakly} propose a method to transfer knowledge from a set of familiar objects to localize new objects in a collection of weakly supervised images. Their method uses semantic information encoded in word vectors for knowledge transfer. 
In contrast, our method uses the similarity between tasks in training and testing and does not rely solely on a given semantic relationship between the familiar and new classes. 
%In a more recent work, Shi et al.~\cite{Shi2017WeaklySO} use class level similarity on whole images, semantic similarity encoded in word hierarchy, and semantic segmentation of source images to transfer knowledge of things and stuff between source and target classes. Their sophisticated hand designed pipeline requires large amount of target images in order to train their final detector. 
Deselaers et al.~\cite{deselaers2012weakly} transfer objectness scores from source classes and incorporate them into unary terms of a conditional random field formulation. %Furthermore, we have no assumption on requiring to know unseen classes in advance.

Our approach inspired by methods that use the meta-learning paradigm for few-shot classification. These methods %use an episodic training scheme that 
simulate the few-shot learning task during the training phase in which the model learns to optimize over a batch of sampled tasks. The meta-learned method is later used to optimize over similar tasks during testing.
Optimization-based methods~\cite{Sachin2017, Finn2017ModelAgnosticMF}, feature and metric learning methods~\cite{vinyals2016matching, snell2017prototypical,sung2018learning}, and memory augmented-based methods~\cite{santoro2016meta} are just a few examples of modern few-shot learning. While our work is inspired by these methods, it is different in the sense that we do not assume strong supervision for the tasks. In relation networks~\cite{sung2018learning} a similarity function is learned between image pairs and used to classify images from unseen classes. We adopt this method to learn the unary and pairwise potential functions in our graphical model.
%=============================================================================
%
\section{Problem description} \label{sec:basics}
We consider a set $\II$ with a binary relation $R$. The elements of the set are called 
{\em images} in our work for simplicity of exposition. 
A {\em relation} $R$ is simply a subset of $\II \times \II$:
\begin{equation}
\label{eq:relation}
R(e, e') = 
\begin{cases}
    +1, & \text{if } (e, e') \in R \text{~(inputs are related)}\\
    -1, & \text{otherwise}.
\end{cases}
\end{equation} 
A {\em bag} is a set of images, %$v_i = \{ e_{ik} \}$,
thus, a subset of $\II$. 
%Note that $e_{ik}$ denotes the $k$-th image in a bag $v_i$. 
We will be concerned with collections of bags, 
$\VV = \{v_1, v_2, \ldots, v_N \}$. We say that a collection 
$\VV = \{v_1, \ldots, v_N\}$ is {\em consistent} if it
is possible to select images, one from each bag, so that they are 
all related in pairs.  These are known as {\em positive} bags.

Given a consistent collection, $\VV$ and an optional additional bag $\bar{v}$ that we designate as \emph{negative}%
%------------
\footnote{There is no point in having more than one negative bag in
a collection
since its purpose is simply to provide a set of images that are
not compatible with the positive bags, in the sense described.}%
%-----------
, the task is to output a {\em selection} of images, namely
an ordered set $O = (e_1, \ldots , e_N)$
where $e_i$ is from positive bag $v_i$, such that the images are pairwise related, $R(e_i, e_j) = 1$, and that not all images are pairwise related to any image in the negative bag, i.e., $\exists e_i \in O$ such that $\forall \bar e \in \bar{v}, R(e_i, \bar e) = -1$.
%
%\subsection{Constructing $R$ from Image Labels}

The situation of most interest is where each of the images $e\in\II$ has a single latent (unknown) label $c_e\in\{c_{\varnothing}\} \cup \mathscr{C}$ where $c_{\varnothing}$ is a {\em background class}
and $\mathscr{C}$ is a set of {\em foreground classes}. Two images
$e_1$ and $e_2$ are related if their labels are the same and belong to a foreground class, i.e., $c_{e_1} = c_{e_2} \in \mathscr{C}$. For example, (cropped) images may be labelled
according to the foreground object they contain. In this case, two images $(e_1, e_2)$, both containing a
``cake'' are related, $R(e_1, e_2) = 1$. Whereas two images $(e_3, e_4)$ that are not of the same foreground class are unrelated, $R(e_3, e_4) = -1$.  In this
case, $R$ is an equivalence relation.

\paragraph{Energy function. }
We pose the problem of finding the common object as finding a selection $O$ that minimizes an energy function. Our energy function is defined as sum of potential functions as follows:
\begin{equation}
\label{eq:gen_mrf}
    E(O \mid \bar v) = \sum_{\substack{e_i,e_j \in O \\ i > j}}
    \psi^{\textrm{P}}_{\theta}(e_i, e_j) + \eta \sum_{e_i \in O} \psi^{\textrm{U}}_{\beta}(e_i \mid \bar v),
\end{equation}
in which $\psi^{\textrm{P}}_{\theta}(\cdot, \cdot)$ and $\psi^{\textrm{U}}_\beta(\cdot \mid \bar v)$ are pairwise and unary potential functions with trained parameters $\theta$ and $\beta$, and hyperparameter $\eta \geq 0$ controls the importance of the unary terms.
Both unary and pairwise potential functions
are learned by neural networks, which will be described in 
Section~\ref{sect:training}.
The pairwise potential function is learned so that it encourages choosing pairs that are related to each other. The unary potential is chosen so it is minimized when its input is not related to the images in the negative bag. In this way, the overall energy is minimized when images in $O$ are related to each other and unrelated to images in the negative bag.

\subsection{Training and Test Splits}
%they are both from a foreground class and their label is similar.
% RIH:  You simply cannot do this!!.  Here you are giving a formal
%definition of the problem, and you use undefined terms such
% as foreground class, and what may be meant by classes being
% similar.  Sloppiness like this is what makes papers incomprehensible.
%We follow an episodic sampling strategy to sample positive and negative bags from the a dataset. For each episode, we first randomly sample $M$ classes out of all the possible classes $\mathscr{C}_{\rm train}$. One of these classes is selected to be the target and the rest are non-target classes. Each positive bag is built by randomly sampling one image from the target class and $B-1$ images from the target and non-target classes so there is at least one image from the target class in the positive bags. For simplicity we assume all positive bags have size $B$. A negative bag is built by sampling $\bar B$ examples from non-target classes. We use notation $\VV' \sim \DD$ to draw a random collection from the dataset $\DD$ using the method described above. 
%
For a dataset $\DD\subseteq\II$, we use the notation $\WW\sim\DD$ to
indicate that a random collection $\WW = (\VV, \bar v)$ is drawn from the
dataset. We define the sampling strategy in the implementation details
for each dataset. During training, algorithms have access to a dataset 
$\DD_{\rm train}$ and corresponding ground-truth relation. We construct the relation
for the training dataset based on a set of foreground classes $\mathscr{C}_{\textrm{train}}$
as described above.

Methods are evaluated on samples from a test dataset $\WW \sim \DD_{\rm test}$.
There are no image in common between the training and test datasets. Moreover,
the set of foreground classes $\mathscr{C}_{\rm test}$ used
for the test dataset is disjoint from the set of foreground classes
used during training, i.e.,~$\mathscr{C}_{\rm test} \cap\, \mathscr{C}_{\rm train} =
\varnothing$. At test time we only know whether a bag is positive or negative
with respect to some foreground class. The ground-truth
(which foreground class is common to the positive bags)
is unknown to the algorithm and is only used for evaluating
performance.

%
%\noindent {\bf Notation.} Let $E$ be a selection over a set of positive bags $\VV$, also called as a selection proposal over set $\VV$. $E_\SSX$ denotes part of selection $E$ which is over the subset of positive bags $\SSX \subseteq \VV$. Each selection is formed by sampling exactly one image from each bag in $\SSX$. %We  drop the subscript when the subset $\SSX$ is clear from the context.
%
%We call two selections \emph{non-overlapping} if they are samples from disjoint bags. Our greedy matching method works by joining and scoring non-overlapping selection proposals from smaller subproblems. The joining operation forms a new selection by concatenating the two selection sequences.% with length equal to the sum of the length of the two selections. 
%Let $\SSX_1$ and $\SSX_2$ be two disjoint sets of bags and $\AA_{\SSX_1}$ and $\AA_{\SSX_2}$ be sets of selections sampled from these sets respectively. The outer product of two sets of selections $\AA_{\SSX_1} \otimes \AA_{\SSX_2}$ is the set of all the possible selections that can be formed by joining selections in $\AA_{\SSX_1}$ with the selections in $\AA_{\SSX_2}$.

%============================================

\section{Learning the potential functions}
\label{sect:training}
\label{sec:potentials}

We now present the method for learning the pairwise and unary potential functions.
The proposed method relies on an algorithm to estimate a similarity measure of an input image pair $(e, e')$. One common approach is to learn an embedding function and use a fixed distance metric to compare the input pairs in the embedded space. In this approach, the learning is used only to determine the embedding function. The relation network~\cite{sung2018learning} extends this by jointly learning the embedding function and a comparator. The network consists of embedding and relation modules. The {\em embedding module} learns a joint feature embedding (into $\R^d$) for the input pair of images $\CC(e, e')$ and the {\em relation module} 
learns a mapping $g: \R^d \rightarrow \R$, mapping the embedded feature to a relation score $r_\phi(e, e') = g(\CC(e, e'))$
where $\phi$ denotes the parameters of the embedding and scoring functions combined.%
\footnote{We adopt the notation used in the relation network paper~\cite{sung2018learning}} 
We adopt the relation module from the Relation Network due to its simplicity and success in few-shot learning. However, any other method which computes the relationship between a pair of images could be used in our method.

\paragraph{Relation network. } As we need to evaluate the relation of many image pairs, we adapt the original relation network architecture~\cite{sung2018learning} in order to make the embedding and scoring functions as computationally efficient as possible. The feature embedding function $\CC(\cdot, \cdot): \II \times \II \rightarrow \R^d$ consists of feature concatenation and a single linear layer with gated activation~\cite{van2016conditional} and skip connections. % shown in Fig~\ref{fig:feature_learning}.  
 Let $f$ and $f'$ be features in $\R^d$ extracted from images $e$
and $e'$ by a CNN {\em feature extraction module} and $[f, f']$ be the concatenation of feature pairs. The embedding function is defined as:
\[
%\label{eq:embedding_module}
    \CC(e, e') =  \tanh(W_1 [f, f'] + b_1)  \sigma(W_2 [f, f'] + b_2) 
    + \frac{f + f'}{2}
\]
where $W_1, W_2 \in \R^{d\times 2d}$ and vectors $b_1, b_2 \in \R^{d}$ are the parameters of the feature embedding module and $\tanh(\cdot)$ and $\sigma(\cdot)$ are hyperbolic tangent and sigmoid activation functions respectively, applied componentwise to vectors in $\R^d$.
%
%\begin{figure}
%    \centering
%    \includegraphics[width=0.6\columnwidth]{feature_learning}
%    \caption{\em \label{fig:feature_learning}Feature Embedding Module $\CC(., .)$. Input feature pairs are embedded into a joint embedding function by a gated activation layer.}
%    \vspace{-1em}
%\end{figure}
%
Then, we use a linear layer to map this features into relation score
\[
r_\phi(e, e') = w^\top\CC(e, e') + b
\]
where $w \in\R^d$ and $b \in \R$. We found in practice that using gated activation in the embedding module improves the performance over a simple ReLU, whereas adding more layers does not affect the performance. We note that the effectiveness of gated activation has also been shown in other work~\cite{activation}. 

%-----------------------------------------------------------------------
\paragraph{Pairwise potentials. }
The pairwise potential function is defined as the negative of the output of the relation module: $\psi^{\textrm{P}}_\theta(e_i, e_j) = -r_\theta(e_i, e_j)$ so it has a lower energy for related pairs. For a sampled collection $\VV$ the episode loss is written as a binary logistic regression loss
\begin{equation*}
\LL^{\textrm{P}} = \frac{1}{N_P} \sum_{(e_i, e_j) \sim \VV} \log{\Big(1+ \exp({-R(e_i, e_j)r_\theta(e_i, e_j)})\Big)}
\end{equation*}
where the sum is over all the pairs in the collection, $N_P$ is the total number of such pairs, and relation $R(., .)$ defined in Eq~\eqref{eq:relation} provides the ground-truth labels.

Note that image pairs are sampled from $\VV$, so that the
loss function reflects prior distributions of image pairs
from consistent image collections.

\paragraph{Unary potentials. } The unary potential $\psi^{\textrm{U}}(e \mid \bar v)$ is constructed by comparing image $e$ with images in the negative bag $\bar v$. Let the vector $u(e, \bar v)$ be the estimated relation between image $e$ and all the images in $\bar v$, that is, $u(e, \bar v)_j = r_\beta(e, \bar e_j)$ where $\bar e_j$ is the $j$-th image in the negative bag and $\beta$
is the (new) set of parameters for the relation network.  By definition, the unary energy for an image $e$ should be high if at least one of the values in $u(e, \bar v)$ is high. In other words, $e$ is related to $\bar v$ if it is related to at least one image in $\bar v$. This suggests the use of ${\rm max}_j(u(e, \bar v)_j)$ as the unary energy potential. However, depending on the class distribution of images in the negative bag, an image $e$ which is not from the common object class could be related to more than just one image from the negative bag. In this case, using the average relation to the few mostly related elements in $u(e, \bar v)$ helps to reduce the noise in the estimation and works better than a simple max operator. This motivates us to use a form of exponential-weighted average of the relations so that higher values get a higher weight
\begin{equation}
    \psi^{\textrm{U}}_{\beta, \nu}(e \mid \bar v) = 
    \frac
    {\sum_{k=1}^{\bar B} u(e, \bar v)_k \exp\big(\nu\, u(e, \bar v)_k\big)}
    {\sum_{k=1}^{\bar{B}} \exp(\nu\, u(e, \bar v)_k)} ~.
\label{eq:unary_function}
\end{equation}
Here, $\bar B$ is the total number of images in the negative bag and $\nu$ is the temperature parameter. Observe that for $\nu = 0$ we have the mean value of $u(e, \bar v)$ and it converges to the max operator as $\nu \rightarrow +\infty$. We let the algorithm learn a balanced temperature value in a data-driven way.

For a sampled collection $\WW = (\VV, \bar v)$, the episode loss for the unary potential is defined as a binary logistic regression loss
\begin{equation}
\label{eq:loss_unary}
    \LL^{\textrm{U}} = \frac{1}{N_U} \sum_{v \in \VV}\sum_{e \in v} \log{\Big(1+\exp(-R(e, \bar v) \psi^{\textrm{U}}_{\beta, \nu}(e, \bar v))\Big)}
\end{equation}
where we use an extended definition of the relation function where $R(e, \bar v) = \max_{\bar e \in \bar v} R(e, \bar e)$ and $N_U$ is the total number of images in all positive bags in the collection. 
Through training, this loss is minimized over choices
of parameters $\beta$ of the relation network, and
the weight parameter $\nu$.
By optimizing this loss, we learn a potential function that has higher value if $e$ is related to one example in the negative bag. Note that in Eq~\eqref{eq:gen_mrf} selection of unary potentials with high values are discouraged.

As before, training samples are chosen from collections
$\WW$ to reflect the prior distributions of related and
unrelated pairs.

Parameters of the unary and pairwise potential functions are learned separately by optimizing the respective loss functions over randomly sampled problems from the training set.
Although both unary and pairwise potential functions use the relation network with an identical architecture, their input class distributions are different, since one is comparing images in positive bags and one is comparing images in positive and negative bags. Thus, sharing their parameters decreases overall performance.

\subsection{Inference} \label{sec:tree}
Finding an optimal selection $O$ that minimizes the energy function defined in Eq~\eqref{eq:gen_mrf} is NP-hard and thus not feasible to compute exactly,
except in small cases.  Loopy belief propagation~\cite{weiss2001optimality}, TRWS~\cite{kolmogorov2006convergent}, and AStar~\cite{bergtholdt2010study}, are among the many algorithms used for approximate energy minimization. We propose an alternative approach specifically designed for solving our optimization problem.

Our approach is designed to decompose the overall problem into smaller subproblems, solve them, and combine their solutions to find a solution to the overall problem. This is based on the observation that a solution to the overall problem will also be a valid solution to any of the subproblems. Let $\VV(p,q) = \{v_p, v_{p+1}, ..., v_q\}$ be a subset of $\VV$. Then, a subproblem refers to finding a set of common object proposals $\BB$ for $\VV(p, q)$ with low energy values; $\BB$ represents
a collection of proposed selections of images from the set of bags $\VV(p,q)$.  The energy value for a selection $O_{p,q} \in \BB$ is defined as sum of all pairwise and unary potentials in the subproblem, similar to how the energy function is defined for the overall problem in Eq~\eqref{eq:gen_mrf}. 

The decomposition method starts at the root (i.e., full problem) and divides the problem into two disjoint subproblems and recursively continues dividing each into two subproblems until each subproblem only contains a single bag $v_i$. If $N = 2^Z$, then this can be represented as a full binary tree\footnote{This is without loss of generality since zero padding could be used if the number of positive bags is not a power of $2$.} where each node represents a subproblem. Let $\NN_i^l$ be the $i$-th node at level $l$. Then root node $\NN_1^Z$ represents the full problem, nodes $\NN_i^l$ at any given level $l$ represent disjoint subproblems of the same size, and the leaf nodes, $\NN_i^0$, at level $0$ of the tree each represent a subproblem with only one positive bag $v_i$.

Each level in the tree maintains a set of partial solutions to the root problem. Computation starts at the lowest level (leaf nodes) where each partial solution is simply one of the images for all images in the bag. At the next level, each node combines the partial solutions from its child nodes and prunes the resulting set to form a new set of partial solutions for its own subproblem, which in turn is used as input to nodes at the next level in the tree and so on until we reach the root node, which is the output for the optimization.
%of this procedure. 
The joining procedure used to combine the partial solutions from two child nodes is described next.

\noindent{\bf Joining:} Node $i$ at level $l$ receives as input solution proposals $\BB^{l-1}_{2i-1}$ and $\BB^{l-1}_{2i}$ from its child nodes $\NN^{l-1}_{2i-1}$ and $\NN^{l-1}_{2i}$. The joining operation simply concatenates every possible selection from the first set with every possible selection in the second set and forms a set of selection proposals $\XX^l_i$ for the subproblem
\begin{equation*}
\XX^l_i = \{[O^l, O^r] ~|~  O^l \in \BB^{l-1}_{2i-1}, O^r \in \BB^{l-1}_{2i}\}
\end{equation*}
where $[\cdot,\cdot]$ concatenates two selection sequences. We denote the joining operation by the Cartesian product notation, i.e., $\XX^l_i = \BB^{l-1}_{2i-1} \times \BB^{l-1}_{2i}$.

\noindent{\bf Pruning:} Since combining the partial solutions from two nodes results in a quadratic increase in the number of partial solutions, the number of potential solutions grows exponentially as we ascend the tree. Also, not all the generated partial solutions contain a common object. Therefore, we use a pruning algorithm $\BB^l_j={\rm prune}(\XX^l_j; k)$ that picks the $k$ selections with the lowest energy values. The energy values for each subproblem can also be efficiently computed from bottom to top. At the lowest level, the energy for each selection is the unary potential from Eq~\eqref{eq:unary_function}, %as
\begin{equation}
E^0_i(O_{i,i}) = \eta \psi^{\textrm{U}}_\beta(e_i \mid \bar v) \quad\forall e_i \in \BB^0_i = v_i,
\end{equation}
Note that selection $O_{i,i} = (e_i)$ consists of only one image. % at this level. 
Starting at the leaves, energy in all nodes can be computed recursively. Let $O \in \XX^l_i$ be formed by joining two selection proposals $O^l \in \BB^{l-1}_{2i-1}$ and $O^r \in \BB^{l-1}_{2i}$. 
The energy function $E^l_i(O)$ can be factored as
\begin{equation}
\label{eq:update_energy}
    E^l_i(O) = E^{l-1}_{2i-1}(O^l) + E^{l-1}_{2i}(O^r) + P(O^l, O^r)
\end{equation}
where $P(\cdot, \cdot)$ is the sum of all pairwise potentials on edges joining the  two subproblems and is computed on the fly. 
\begin{algorithm}[t]
\DontPrintSemicolon
\KwIn{$\VV = \{v_1, ..., v_N\}$, $\bar v$, and $N=2^Z$.}
\KwOut{Selection $O = (e_1, \dots, e_N)$}
$\BB_i^0 = v_i ~~~\forall i \in [1, \dots N]$\;
$E_i^0(O_{i,i}) = \eta \psi^{\textrm{U}}_\beta(e_i \mid \bar v) ~~~\forall e_i \in \BB_0^i,~i \in [1, \dots, N]$\;
\For{$l \gets 1$ \textbf{to} $Z$}{
  \For{$i \gets 1$ \textbf{to} $2^{Z-l}$}{
    $\XX_i^l \gets \BB^{l-1}_{2i-1} \times \BB^{l-1}_{2i}$ (joining)\;
    Compute $\XX_i^l$ Energies According to Eq~\eqref{eq:update_energy}\;
    $\BB_i^l \gets \texttt{prune}(\XX_i^l; k)$ (pruning)\;
  }
}
\Return{$O \in \BB^Z_1$ with the minimum energy}\;
\caption{\label{alg:greedy} Greedy Optimization Algorithm}
\end{algorithm}
Algorithm~\ref{alg:greedy} summarizes the method. A good value of $k$ in the pruning method depends on the ambiguity of the task. It is possible to construct an adversarial example that needs \emph{all} possible proposals at the root node to find the optimal solution. However, in practice, we found that $k$ does not need to be large to achieve good performance.
%comparable to other energy minimization methods.
Importantly, unlike other methods, this algorithm does not necessarily compute all of the pairwise potentials. For example, if an object class only appears in a small subproblem, the images of that class will get removed by nodes whose subproblem size is large enough. Thus, in the next level of the tree, the pairwise potentials between those images and other images is no longer required. In general, the number of pairwise potentials computed depends on both the value of $k$ and the dataset. We observed that only a small fraction of the total pairwise potentials
were required in our experiments.
%In the co-localization experiments from Section \ref{sec:exp_co_loc} on average only $15\%$ of the total pairwise potentials were evaluated.

\section{Experiments} \label{sec:experiment}
We evaluate the proposed algorithm on few-shot common object recognition and co-localization tasks. For each task, we first pre-train a CNN feature extractor module to perform classification on the seen categories from the training dataset. We then use the learned CNN to compute a feature descriptor of each image. This ensures a consistent image representation for all methods under consideration.
%The details of feature extractor modules will explained for each task in it subsection. 

For learning pairwise and unary potentials, stochastic gradient descent with gradual learning rate decay schedule is used.
%to minimize the loss function.
%in Eq~\eqref{eq:total_loss}. 
The complete framework (``Ours'' in the tables) uses greedy optimization method described in Algorithm \ref{alg:greedy}. The optimal value of $\eta$ in Eq~\eqref{eq:gen_mrf} is found using grid search. In all experiments, a maximum of $k = 300$ top selection proposals are kept in the greedy algorithm.

All experiments are done on a single Nvidia GTX 2080 GPU and 4GHz AMD Ryzen Threadripper 1920X CPU with 12 Cores\footnote{The code is publicly available \href{https://github.com/haamoon/finding_common_object}{here}.}.
%
%We will make our code and models available to the public upon acceptance. 
%In the next sections, we first review the baseline methods and then present the results for each task.
%
\subsection{Baseline Methods}
We compare the greedy optimization algorithm to AStar~\cite{bergtholdt2010study} which is used for object co-segmentation~\cite{vicente2011object} and the faster TRWS~\cite{kolmogorov2006convergent} which is used for inference on MIL problems~\cite{Deselaers2010, deselaers2012weakly}. We use a highly efficient parallel implementation of these algorithms~\cite{opengm-library}. The proposed method is compared to SVM based and attention based MIL baselines described below.

\noindent{\bf SVM based MIL. } We report the results of the three well-known approaches: MI-SVM~\cite{Hoffman2015DetectorDI}, mi-SVM~\cite{andrews2003support} and sbMIL~\cite{Bunescu:2007:MIL:1273496.1273510} using publicly available source code~\cite{Doran:2014:TEA:2666867.2666935}. The sbMIL method is specially designed to deal with sparse positive bags. 
%We report the results of Hoffman~\emph{et al.}~\cite{Hoffman2015DetectorDI} in which an embedding function is learned in two steps: first, bag labels are used to pre-train the parameters of a CNN. 2) next, fully labeled training data is used to fine-tune the pre-trained parameters. In our experiments, we skip the first step since the problem size is too small for pre-training. 
The RBF and linear kernel are chosen as they work better on few-shot common object recognition and co-localization respectively. Grid search is performed in order to select the hyperparameters.

\noindent{\bf Attention based deep MIL. }
Along with the SVM based methods, the results of the more recent attention based deep learning MIL method~\cite{pmlr-v80-ilse18a} (ATNMIL) is presented on our benchmarks. After training the model, we select the image proposal with the maximum attention weight from each positive bag.

\subsection{Few-shot Common Object Recognition}
\label{sec:exp_few_shot_common_obj_rec}
In this task we make use of the \textit{mini}ImageNet dataset~\cite{vinyals2016matching}. The advantage of \textit{mini}ImageNet is that we can compare many different design choices without requiring large scale training and performance evaluations. The dataset contains $60,000$ images of size $84\!\times\!84$ from $100$ classes. We experiment on the standard split for this task of $64$, $16$ and $20$ classes for training, validation and testing, respectively~\cite{Sachin2017}. 

For the CNN feature extractor module, a Wide Residual Network~(WRN)~\cite{BMVC2016_87} with depth $28$ and width factor $10$ is pre-trained on the training split.
% using $500$ images per class while keeping the rest for validation. We use similar configuration as used in \cite{BMVC2016_87} for CIFAR10 dataset to train the WRN.
The $d=640$ dimensional output of global average pooling layer of the pre-trained network is provided as input to all the methods. %corresponding to each image as input . %in all of our experiments on \textit{mini}ImageNet.

To construct bags, we first randomly select $M$ classes out of all the possible classes $\mathscr{C}$. One of these is selected to be the target and the rest are considered non-target classes. Then, each positive bag is constructed by randomly sampling one image from the target class and $B-1$ images from the target and non-target classes. 
The negative bag is built by sampling $\bar B$ examples from non-target classes. For output selection $O$, we measure the {\em success rate} which is equal to the percentage 
of $e \in O$ that belong to the target class. We compute the expected value of success rate for $1000$ randomly sampled problems and report the mean and $95$\% confidence interval of the evaluation metric.

We vary the number of bags as well as their sizes. We select the number of positive bags $N \in \{4,8,16\}$, the size of each positive bag $B \in \{5,10\}$, and the size of negative bag $\bar B \in \{10,20\}$. The number of classes $M$ to sample from in each episode changes the difficulty of the task. We randomly choose $M$ between $5$ and $15$ when $B = 5$, and between $10$ and $20$ when $B = 10$ for each problem.
%Lower values of $M$ make the problem more ambiguous by increasing the chance of generating other common objects in subproblems. On the other hand, it increases the importance of the negative bag by increasing the chance of having more samples from each non-target class. 
%To generate problems with different difficulty levels, at each iteration, 

The results in Table~\ref{table:miniimagenet} show our method outperforms ATNMIL and SVM based approaches for all versions of the problem. To test the importance of learning the unary and pairwise potentials, we construct a baseline that uses cosine similarity to compute the relation between pairs\footnote{We also use negative of Euclidean distance measure for the relation but it shows inferior performance.} while keeping the rest of the algorithm identical. The performance gap between our method and the baseline shows that the relation learning method, apart from structured inference formulation, plays an important role in boosting the performance. 

\begin{figure}[t]
\begin{center}
\includegraphics[width=0.33\textwidth, trim=.0 1.05cm 2.2cm 0.6cm, clip]{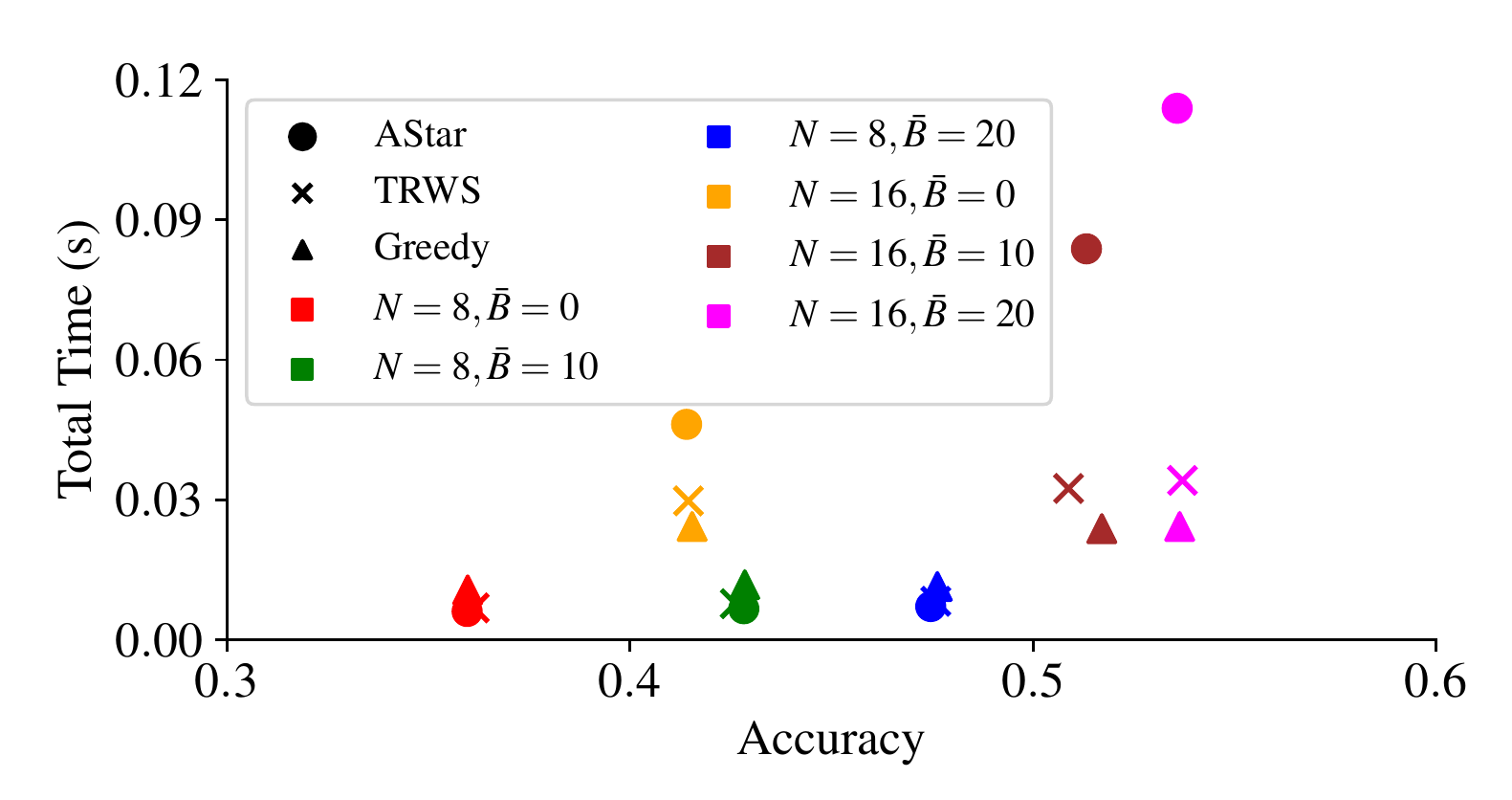}
\caption{\label{fig:runtime}\em Average runtime vs. accuracy of different inference algorithms on \textit{mini}ImageNet for $N \in \{8,16\}$, $\bar B \in \{0, 10, 20 \}$, and $B=10$. 
%Circle, cross, and triangle correspond to AStar, TRWS and Greedy inference, respectively. 
Each setting is shown with a distinct color.
}
\end{center}
\vspace{-0.9cm}
\end{figure}

Average total (potentials computation and inference) runtime versus accuracy plot of different energy minimization methods on different settings is shown in Figure \ref{fig:runtime}. Even on this small scale problem, the greedy optimization is faster on average while its accuracy is on par with other inference methods. See 
%Table~\ref{table:miniimagenet_energy_minimization} in 
the supplementary material for complete numerical results.

\begin{table*}
\begin{minipage}[t]{.67\textwidth}
\centering
\resizebox{1.0\textwidth}{!}{%  
\begin{tabular}{c c || c c | c c | c c} 
& $N$ & \multicolumn{2}{c}{\Large{$4$}} & \multicolumn{2}{c}{\Large{$8$}} & \multicolumn{2}{c}{\Large{$16$}} \\
& $\bar B$ & $10$ & $20$ & $10$ & $20$ & $10$ & $20$ \\
\hlineB{3}
\multirow{6}{*}{\rotatebox[origin=c]{90}{Bag Size = $5$}}
& Ours & ${\bf 63.83\pm 1.49}$ & ${\bf 65.48\pm1.47}$  & ${\bf 72.49\pm 0.98}$ & ${\bf 73.99\pm0.96}$ & ${\bf 78.60\pm 0.64}$ & ${\bf 79.93\pm0.62}$ \\
& Baseline & $60.88\pm1.51$ & $63.83\pm1.49$  & $64.46\pm1.05$ & $68.08\pm1.02$ & $66.78\pm0.73$ & $70.39\pm0.77$ \\
& MI-SVM~\cite{Hoffman2015DetectorDI} & $56.25\pm 1.54$ & $59.03\pm 1.52$  & $62.75\pm 1.06$ & $63.76\pm 1.05$ & $67.91\pm 0.72$ & $73.33\pm 0.69$ \\
& sbMIL~\cite{Bunescu:2007:MIL:1273496.1273510} & $54.55\pm 1.54$ & $59.93\pm 1.52$  & $58.25\pm 1.08$ & $64.68\pm 1.05$ & $61.35\pm 0.75$ & $65.55\pm 0.74$ \\
& mi-SVM~\cite{andrews2003support} & $54.23\pm 1.54$ & $59.43\pm 1.52$  & $60.43\pm 1.07$ & $66.08\pm 1.04$ & $64.49\pm 0.74$ & $69.69\pm 0.71$ \\
& ATNMIL~\cite{pmlr-v80-ilse18a} & $50.35\pm 1.55$ & $60.33\pm 1.52$ & $56.05\pm 1.09$ & $63.29\pm 1.06$ & $58.97\pm 0.76$ & $67.26\pm 0.73$ \\
\hline
\multirow{6}{*}{\rotatebox[origin=c]{90}{Bag Size = 10}}
& Ours & ${\bf 37.42\pm1.50}$ & $38.50\pm1.51$ & ${\bf 42.85\pm1.08}$ & ${\bf 47.63\pm1.09}$ & ${\bf 51.70\pm0.77}$ & ${\bf 53.63\pm0.77}$ \\
& Baseline & $35.73\pm1.49$ & ${\bf 40.40\pm1.52}$ & $38.01\pm1.06$ & $43.95\pm1.09$ & $41.08\pm0.76$ & $47.83\pm0.77$ \\
& MI-SVM~\cite{Hoffman2015DetectorDI} & $29.53\pm 1.41$ & $35.05\pm 1.48$ & $35.25\pm 1.05$ & $39.94\pm 1.07$ & $41.21\pm 0.76$ & $46.63\pm 0.77$ \\
& sbMIL~\cite{Bunescu:2007:MIL:1273496.1273510} & $31.55\pm 1.44$ & $31.50\pm 1.44$ & $34.10\pm 1.04$ & $39.86\pm 1.07$ & $28.80\pm 0.70$ & $43.63\pm 0.77$ \\
& mi-SVM~\cite{andrews2003support} & $31.55\pm 1.44$ & $35.33\pm 1.48$ & $34.10\pm 1.04$ & $39.86\pm 1.07$ & $39.48\pm 0.76$ & $45.16\pm 0.77$ \\
& ATNMIL~\cite{pmlr-v80-ilse18a} & $26.58\pm 1.37$ & $33.10\pm 1.46$ & $28.48\pm 0.99$ & $35.11\pm 1.05$ & $31.56\pm 0.72$ & $38.14\pm 0.75$ \\
\end{tabular}}
\caption{\em \label{table:miniimagenet}\small{Success rate on \textit{mini}ImageNet for different positive bags $N$, and total number of negative images $\bar B$. The first and the second part of the table show the results for bag size $5$ and $10$ respectively.}}
\end{minipage}\quad
\begin{minipage}[t]{.32\textwidth}
\centering
\resizebox{1.0\textwidth}{!}{%
\begin{tabular}{c||c | c}
Method & COCO & ImageNet\\
\hlineB{3}
MI-SVM~\cite{Hoffman2015DetectorDI} & $60.74 \pm 1.07$ & $49.44 \pm 1.10$\\
ATNMIL~\cite{pmlr-v80-ilse18a} & $60.00 \pm 1.07$ & $49.35 \pm 1.10$\\
Ours & $\mathbf{65.34 \pm 1.04}$ & $\mathbf{55.18 \pm 1.09}$\\
\hline
TRWS~\cite{kolmogorov2006convergent} & $65.04 \pm 1.05$ & $54.20 \pm 1.09$\\
AStar~\cite{bergtholdt2010study} & $64.99 \pm 1.05$ & $54.23 \pm 1.09$\\
\hline
Unary Only & $59.24 \pm 1.08$ & $50.29 \pm 1.10$\\
TRWS Pairwise Only & $64.53 \pm 1.05$ & $52.95 \pm 1.09$\\
AStar Pairwise Only & $64.54 \pm 1.05$ & $52.89 \pm 1.09$\\
Ours Pairwise Only & $64.65 \pm 1.05$ & $53.00 \pm 1.10$\\
\end{tabular}}
\vspace{.45cm}
\caption{\em \label{table:coco}\small{CorLoc(\%) on COCO and ImageNet with $8$ positive and $8$ negative images.}}
\end{minipage}
\vspace{-0.7cm}
\end{table*}

%-----------------------------------------------------------

\subsection{Co-Localization}
\label{sec:exp_co_loc}
We evaluate on the co-localization problem to illustrate the benefits of the methods discussed in the paper on a real world and large scale dataset. In this task, we train the algorithm on a split of COCO 2017~\cite{lin2014microsoft} dataset with $63$ seen classes and evaluate on the remaining $17$ unseen classes. The resulting dataset contains $111,085$ and $8,245$ images in the training and test set respectively. To evaluate the performance of the trained algorithm on a larger set of unseen classes we also test on validation set of ILSVRC2013 detection~\cite{ILSVRC15}. This dataset has originally $200$ classes but only $148$ classes do not have overlap with the classes that were used for training. The final dataset, after removing coco seen classes, contains $12,544$ images from $148$ unseen classes.
The dataset creation method is explained in the supplementary material in more detail.

For the CNN feature extractor module, we pre-train a Faster-RCNN detector~\cite{girshick2015fast} with ResNet-50~\cite{he2016deep} backbone on the COCO training dataset which has only seen classes. For each image, region proposals with the highest objectness scores are kept. The output of the second stage feature extractor is used in all methods.

For this task, each bag is constructed by extracting top $B=300$ region proposals from one image and a selection $O$ represents one bounding box from each image. To select images of each problem, we first randomly select one class as the target. Then, $N$ images which have at least one object from the target class are sampled as positive bags. The negative bag is composed of images which do not contain the target class. 
%We extract $B=300$ region proposals out of each image. 
The success rate metric used in few-shot common object detection is used to evaluate the performance of different algorithms. A region proposal is considered successful if it has IoU overlap greater than $0.5$ with the ground-truth target bounding box. Note that for the co-localization task, this metric is equivalent to class agnostic CorLoc~\cite{deselaers2010localizing} measure which is widely used for localization problem evaluation~\cite{46626,Shi2017WeaklySO,bilen2015weakly,cinbis2017weakly}.

\begin{figure*}[ht]
\begin{center}
\includegraphics[width=0.87\textwidth, trim=.0 12.0cm 0.0cm 0.0cm, clip]{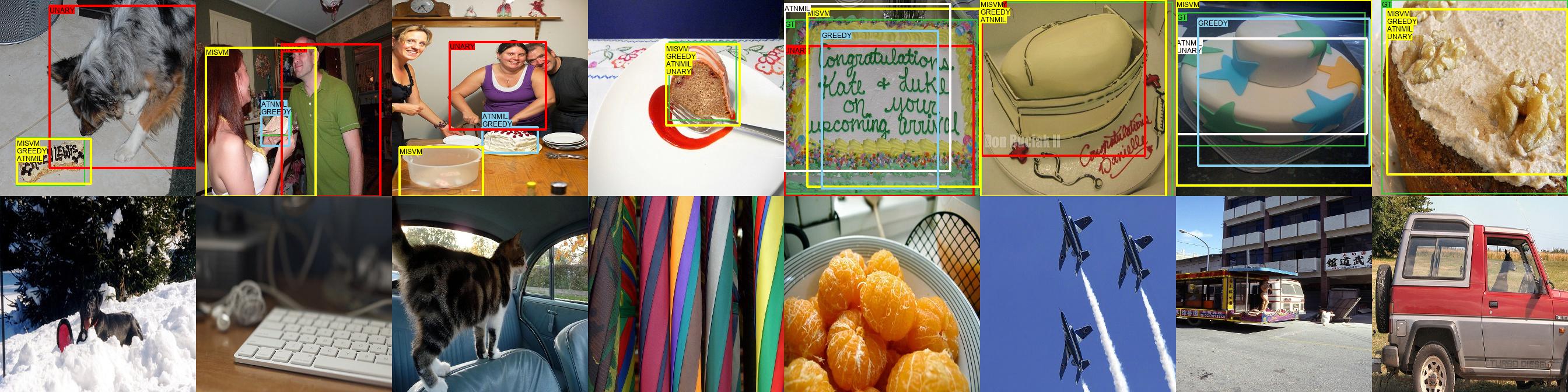} \\
\includegraphics[width=0.87\textwidth, trim=.0 12.0cm 0.0cm 0.0cm, clip]{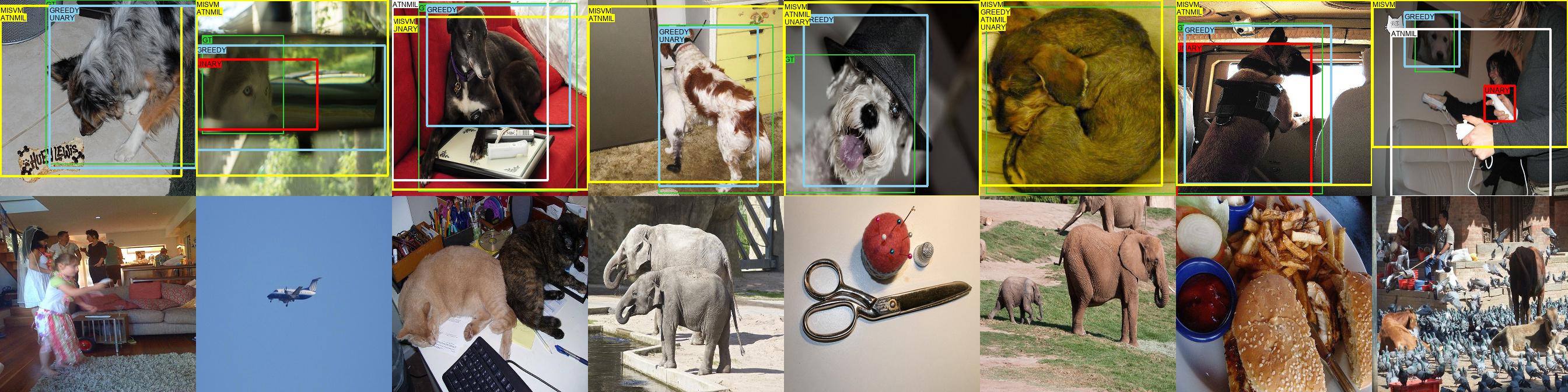} \\
\includegraphics[width=0.87\textwidth, trim=.0 12.0cm 0.0cm 0.0cm, clip]{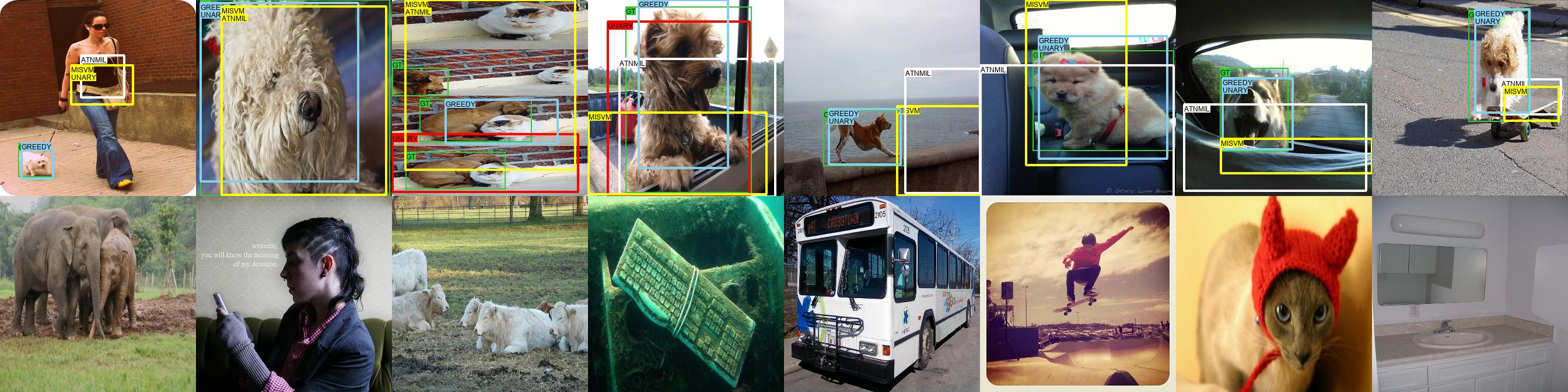} \\
\includegraphics[width=0.87\textwidth, trim=.0 12.0cm 0.0cm 0.0cm, clip]{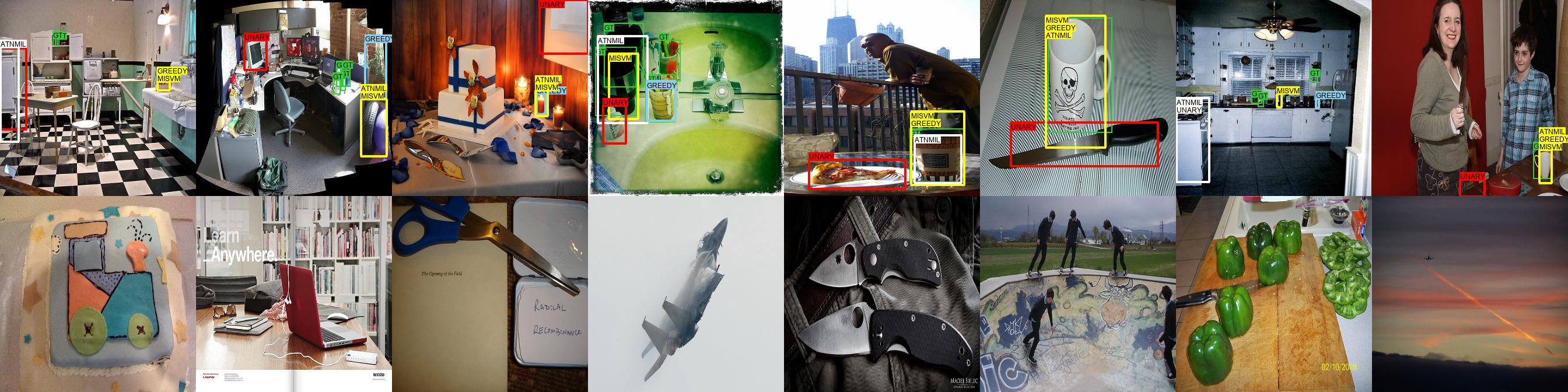}
\caption{\em Qualitative results on COCO dataset. Each row shows positive bags of a sampled collection. Negative bags are not shown. Note that the first image in the first two rows are identical but the target object is different. Last row shows a failure case of our algorithm. While cup is the target object, our method finds plant in the second image. This might be due to the fact that pot (which has visual similarities to cup) and plant are labelled as one class in the training dataset. Note that ``dog'', ``cake'' and ``cup'' are samples from unseen classes. Selected regions are tagged with method names. Ground-truth target bounding box is shown in green with tag ``GT''.} \label{fig:coco_qualitative}
\end{center}
\vspace{-0.9cm}
\end{figure*}

Table~\ref{table:coco} illustrates the quantitative results on COCO and ImageNet datasets with $8$ positive and $8$ negative images\footnote{We skip the results for sbMIL and mi-SVM as they showed similar or inferior results to MI-SVM.}. Our method works considerably better than other strong MIL baselines. Qualitative results of our method compared with other MIL based approaches are illustrated in Figure~\ref{fig:coco_qualitative}. Our method selects the correct object even when the target object is not salient. More qualitative results are presented in the supplementary material.

\begin{figure}[htbp]
\begin{center}
\includegraphics[width=0.27\textwidth, trim=.0 1.1cm 0.0cm 0.6cm, clip]{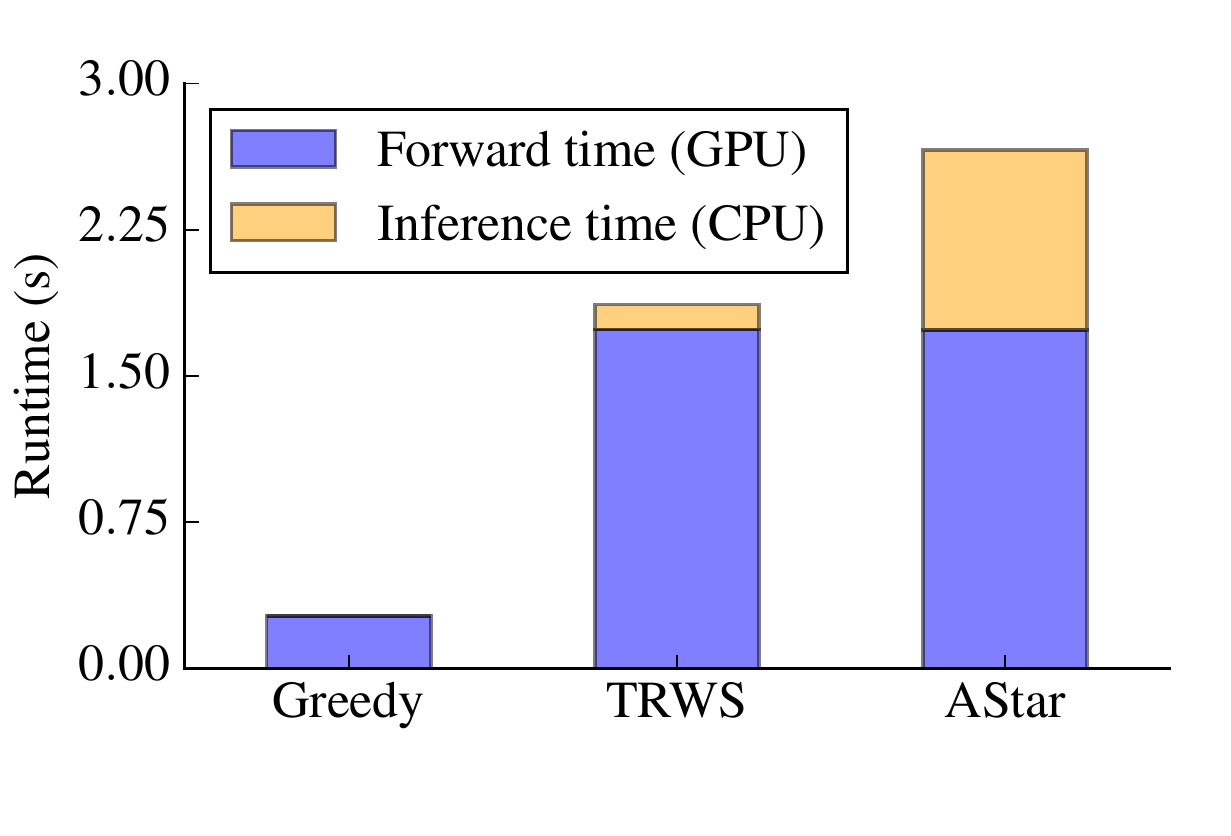}
\caption{\em Forward and inference time (in sec.) on COCO.} \label{fig:runtime_coco}
\end{center}
\vspace{-1.0cm}
\end{figure}

To see the effect of unary and pairwise potentials separately, we provide results for two new variants for structured inference based methods: (i)  Unary Only: where the common object proposal in each bag is selected using only the information in negative bags without seeing the elements in other bags, and (ii) Pairwise Only: where the negative bag information is ignored in each problem. The results show that the pairwise potentials contribute more to the final results. This is not surprising since negative images only help when they contain an object which is also appearing in positive images which, given the number of classes we are sampling from, has a low chance. Interestingly, by using the learned unary potentials alone we could get comparable results to the MIL baselines.
The results in Table~\ref{table:coco} show that different inference algorithms have very similar performance. However, as it is shown in  Figure~\ref{fig:runtime_coco}, the greedy optimization algorithm is much faster. Note that our method requires to compute only 15\% out of all pairwise potentials in average.
%In order to compute the pairwise potentials for dense graphical models, we require to forward different combination of pairs multiple times on a single GPU. We forward them in a number of batches which take the maximum capacity of the GPU memory (11GB). 
One may argue that the pairs can be forwarded on multiple GPUs in parallel and this reduces the forward time. However, our greedy inference method can also take advantage of multiple GPUs since the nodes at each level are data independent.
%
%\begin{table}[h]
%\begin{center}
%\resizebox{0.4\linewidth}{!}{%
%\begin{tabular}{c||c}
%Method & Accuracy ($\%$) \\
%\hlineB{3}
%No Unary & $64.48\pm1.47$ \\
%MEAN & $70.23 \pm 1.00$ \\
%MAX & $71.76 \pm 0.99$ \\
%SOFTMAX & $72.49 \pm 0.98$
%\end{tabular}}
%\caption{\em \label{table:mini_unary}\small{Comparison of different unary potential %functions.}}
%\end{center}
%\vspace{-.5cm}
%\end{table}
%
\begin{table}[h]
\begin{center}
\resizebox{1.0\linewidth}{!}{%
\begin{tabular}{c||c c c c}
Method  & No Unary & MEAN & MAX & SOFTMAX \\
\hlineB{3}
Accuracy ($\%$) & $64.48\pm1.47$ & $70.23 \pm 1.00$ & $71.76 \pm 0.99$ & $\mathbf{72.49 \pm 0.98}$
\end{tabular}}
\caption{\em \label{table:mini_unary}\small{Comparison of different unary potential functions in \textit{mini}ImageNet experiment with $N=8$, $B=5$ and $\bar{B}=10$.}}
\end{center}
\vspace{-.9cm}
\end{table}

\begin{table}[h]
\begin{center}
\resizebox{0.5\linewidth}{!}{%
\begin{tabular}{c||c}
Method & Accuracy ($\%$) \\
\hlineB{3}
adaResNet \cite{munkhdalai2018rapid} & $56.88 \pm 0.62$ \\
SNAIL \cite{mishra2018a} & $55.71 \pm 0.99$ \\
Gidaris et~al. \cite{gidaris2018dynamic} & $55.45 \pm 0.89$ \\
TADAM \cite{oreshkin2018tadam} & $58.50 \pm 0.30$ \\
Qiao et~al. \cite{Act2Param} &  $\mathbf{59.60 \pm 0.41}$ \\
\hline
$\text{Ours-ReLU}$ & $56.43 \pm 0.79$ \\
$\text{Ours-Gated}$ & $57.80 \pm 0.77$
\end{tabular}}
\caption{\em \label{table:fewshot}\small{5-way, 1-shot, classification accuracy with $95\%$ 
confidence interval on \textit{mini}ImageNet test set.}}
\end{center}
\vspace{-.9cm}
\end{table}
\subsection{Ablation Study}
In order to evaluate the effectiveness of our proposed unary potential function, we devise the following experiment. In the few-shot common object recognition task with $N=8$ positive bags, $\bar B=10$ negative images, and $B=5$, we train the unary potentials with four different settings: (1) SOFTMAX: Unary potential function with the learned $\nu$ described in Section \ref{sec:potentials}, (2) MAX: $\nu \rightarrow +\infty$, (3) MEAN: $\nu = 0$, and (4) No Unary: the model without using negative bag information. The pairwise potential function is kept identical in all the methods. The performance of our methods on the described settings are presented in Table \ref{table:mini_unary}. The results show the superiority of the learned weighted similarity to other strategies.

Next, we evaluate the quality of the learned pairwise relations $r(e,e')$ by using them for the task of one-shot image recognition~\cite{vinyals2016matching} on \textit{mini}ImageNet and compare it to the other state-of-the-art methods. In each episode of a one-shot $5$-way problem, $5$ classes are randomly chosen from the set of possible classes and one image is sampled from each class. This mini-training set is used to predict the label of a new query image which is sampled from one of the $5$ classes. The performance is the accuracy of the method to predict the correct label averaged over many sampled episodes. All of these models are trained with a variant of deep residual networks \cite{BMVC2016_87, he2016deep}. Note that unlike other methods, the model in \cite{Act2Param} is trained on validation+training meta-sets.

At each episode, we use the learned relation function to score the similarity between the query image and all the images in the mini training set. The predicted label for the query image is simply the label of the image in mini training set which has the highest relation value to the query image. We compute the accuracy of the predictions of our pairwise potentials on test classes of \textit{mini}ImageNet and compare it with current state-of-the-art few-shot methods in Table \ref{table:fewshot}. We also provide comparison of gated activation function and a simplified ReLU activation in our architecture. Although our method is not trained directly for the task of one-shot learning, it achieves competitive results to the previous methods which are specifically trained for the task. Also, the results show the advantage of using gated activation over ReLU.

\section{Conclusion}
We introduce a method for learning to find images of a common object category across few bags of images which is constructed by learning unary and pairwise terms in an structured output prediction framework. Moreover, we propose an inference algorithm that uses the structure of the problem to solve the task at hand without requiring computation of all pairwise terms. Our experiments on two challenging tasks in the low data regime illustrate the advantage of our knowledge transfer method to several MIL weakly supervised algorithms. In addition, our inference algorithm performs comparable to the well-known structured inference algorithms for this task while being faster.
%\clearpage
%{\small
%\bibliographystyle{unsrt}
\bibliographystyle{ieee_fullname}
\bibliography{egbib}
%}

\iftrue
\clearpage
\onecolumn
\appendix
 \begin{center}
     {\bf \Large Appendix} \\ \vspace{.3cm}
     {\bf \large Learning to Find Common Objects Across Few Image Collections}\\\vspace{.2cm}
     %{\large Paper \#$3345$}
 \end{center}
 
\section{Co-Localization: COCO Dataset Creation and Faster-RCNN Training}
COCO dataset has $80$ classes in total. We take the same $17$ unseen classes which is used in zero-shot object detection paper~\citeRef{bansal2018zero} and keep remaining $63$ classes for training. The training set is constructed using the images in COCO 2017 train set which contain at least one object from the seen classes. The COCO test set, is built by combining the unused images of the train set and images in COCO validation set which contain at least one object from the unseen classes. Similar to~\citeRef{bansal2018zero}, to avoid training the network to classify unseen objects as background, we remove objects from unseen classes from the training images using their ground-truth segmentation masks.

We use Tensorflow object-detection API for pre-training the Faster-RCNN feature extraction module~\citeRef{huang2017speed}. To speed up pre-training, training images are resized down to $336\!\times\!336$ pixels and ResNet-$50$~\citeRef{he2016deep_supp} is used as the backbone feature extractor. %The ResNet weights are initialized with ImageNet pretrained parameters for classification task. 
All layer weights are initialized with variance scaling initialization~\citeRef{glorot2010understanding} and biases are set to zero initially. An additional linear layer which maps the $2048$ dimensional output of second stage feature extractor to a $d=640$ dimensional feature vector is added to the network. We did this to have the dimension of the feature space the same as few-shot common object recognition experiment. We pre-train the feature extractor on four GPUs with batch size of $12$ for $600k$ iterations. The $d=640$ dimensional features are used as input to all of the methods in our experiments.
%We experimented with different number of bag sizes for MIL based methods and found that $B=100$ has slightly better performance. Structured inference based methods, on the other hand, have better performance when the number of proposals is set to $300$.  The criteria used for selecting the top proposals is the class agnostic objectness score of our pretrained Faster-RCNN model.

\section{Hyperparameter Tuning}
In the few-shot common object recognition task, we use grid search on the validation set to tune the hyperparameters of all the methods. To ensure that the structured inference methods optimize the same objective function, we find $\eta$ for the TRWS method and use the same value in AStar and greedy energy functions. For the few-shot common object recognition task value of $\eta$ is shown in Table~\ref{table:miniimagenet_energy_minimization} for each setting. 

In the Co-Localization experiments, the results of the best performing hyperparameters is reported for all the methods. $\eta=0.5$ and $\eta=0.7$ is used in COCO and ImageNet experiments respectively.

\section{Structured Inference Methods Comparison}
The numerical results which are used to generate Figure~\ref{fig:runtime} of the paper are shown in Table~\ref{table:miniimagenet_energy_minimization}. The success rate of the greedy method is on par with the other inference algorithms. From the optimization point of view it is also important to see the mean energy value for the top selection of each method. These results are shown in Table~\ref{table:miniimagenet_energy} and Table~\ref{table:coco_energy} for few-shot common object recognition and co-localization experiments respectively. While AStar and TRWS achieve lower energy values for this problems, the success rate of the methods are comparable. This suggests that finding an approximate solution for the minimization problem is sufficient for achieving high success rate.
\begin{table}[h]
\begin{center}
\resizebox{1.0\textwidth}{!}{%
\begin{tabular}{c c || c c c | c c c | c c c} 
& $N$ & \multicolumn{3}{c}{\Large{$4$}} & \multicolumn{3}{c}{\Large{$8$}} & \multicolumn{3}{c}{\Large{$16$}} \\
& $\bar B$ & $0$ & $10$ & $20$ & $0$ & $10$ & $20$ & $0$ & $10$ & $20$ \\
\hlineB{3}
\multirow{3}{*}{\rotatebox[origin=c]{90}{B = $5$}}
& TRWS & $54.55 \pm 1.54 (0.0)$ & $63.78 \pm 1.49 (0.5)$ & $65.43\pm 1.47 (0.6)$  & $64.55 \pm 1.05 (0.0)$ & $72.60 \pm 0.98 (0.8)$ & $73.80\pm 0.96 (1.2)$ & $70.29\pm 0.71 (0.0)$ & $78.71\pm 0.63 (1.6)$ & $80.08\pm 0.62 (1.9)$ \\
& ASTAR & $54.55 \pm 1.54  (0.0)$ & $63.82\pm 1.49 (0.5)$ & $65.48 \pm 1.47 (0.6)$  & $64.48\pm 1.05 (0.0)$ & $72.49 \pm 0.98 (0.8)$ & $73.99 \pm 0.96 (1.2)$ & $69.91\pm 0.71 (0.0)$ & $78.49\pm 0.64 (1.6)$ & $80.03\pm 0.62 (1.9)$ \\
& Greedy(Ours) & $54.55\pm1.54 (0.0)$ & $63.83\pm 1.49 (0.5)$ & $65.48\pm1.47 (0.6)$  & $64.48\pm1.05 (0.0)$ & $72.49\pm 0.98 (0.8)$ & $73.99\pm0.96 (1.2)$ & $69.67\pm0.71 (0.0)$ & $78.60\pm 0.64 (1.6)$ & $79.93\pm0.62 (1.9)$ \\
\hline
\multirow{3}{*}{\rotatebox[origin=c]{90}{B = 10}}
& TRWS & $29.40\pm 1.41(0.0)$ & $37.15\pm 1.50 (0.5)$ & $38.50\pm 1.51 (0.7)$  & $36.14\pm 1.05 (0.0)$ & $42.61\pm 1.08 (0.9)$ & $47.59\pm 1.09 (1.1)$ & $41.45\pm 0.76 (0.0)$ & $50.88\pm 0.77 (1.5)$ & $53.71\pm 0.77 (2.3)$ \\
& ASTAR & $29.20 \pm  1.41(0.0)$ & $37.43\pm 1.50 (0.5)$ & $38.50\pm 1.51 (0.7)$  & $35.96\pm 1.05 (0.0)$ & $42.83\pm 1.08 (0.9)$ & $47.46\pm 1.09 (1.1)$ & $41.41\pm 0.76 (0.0)$ & $51.32\pm 0.77 (1.5)$ & $53.57\pm 0.77 (2.3)$ \\
& Greedy(Ours) & $29.20\pm1.41 (0.0)$ & $37.42\pm1.50 (0.5)$ & $38.50\pm1.51 (0.7)$ & $35.98\pm1.05 (0.0)$ & $42.85\pm1.08 (0.9)$ & $47.63\pm1.09 (1.1)$ & $41.54\pm0.76 (0.0)$ & $51.70\pm0.77 (1.5)$ & $53.63\pm0.77 (2.3)$ \\
\end{tabular}}
\caption{\em \label{table:miniimagenet_energy_minimization}\small{Success rate of different energy minimization algorithms on \textit{mini}ImageNet. These numbers were used to generate Figure~\ref{fig:runtime} in the paper. Value of the parameter $\eta$ is shown in the parenthesis for each experiment. See section~\ref{sec:exp_few_shot_common_obj_rec} and Table~\ref{table:miniimagenet} for the detailed problem setup.}}
\end{center}
\end{table}
 
\begin{table}[h]
\begin{center}
\resizebox{1.0\textwidth}{!}{%
\begin{tabular}{c c || c c c | c c c | c c c} 
& $N$ & \multicolumn{3}{c}{\Large{$4$}} & \multicolumn{3}{c}{\Large{$8$}} & \multicolumn{3}{c}{\Large{$16$}} \\
& $\bar B$ & $0$ & $10$ & $20$ & $0$ & $10$ & $20$ & $0$ & $10$ & $20$ \\
\hlineB{3}
\multirow{3}{*}{\rotatebox[origin=c]{90}{B = $5$}}
& TRWS & $2.929179$ & $-4.416873$ & $-4.842334$ & $18.300657$ & $-4.425953$ & $-12.602217$ & $86.034355$ & $-6.873013$ & $-10.020649$ \\
& ASTAR & $2.908970$ & $-4.429455$ & $-4.851543$ & $18.192284$ & $-4.529052$ & $-12.666497$ & $85.560267$ & $-7.277377$ & $-10.398633$ \\
& Greedy & $2.908970$ & $-4.429455$ & $-4.851543$  & $18.192282$ & $-4.529052$ & $-12.666499$ & $86.692482$ & $-6.909996$ & $-10.002609$ \\
\hline
\multirow{3}{*}{\rotatebox[origin=c]{90}{B = 10}}
& TRWS & $0.515563$ & $-6.576048$ & $-8.300273$ & $8.749933$ & $-15.959289$ & $-17.238385$ & $53.324193$ & $-28.602048$ & $-59.609459$ \\
& ASTAR & $0.502832$ & $-6.597286$ & $-8.315386$ & $8.675015$ & $-16.079914$ & $-17.404502$ & $52.819455$ & $-29.388606$ & $-60.499036$ \\
& Greedy & $0.502832$ & $-6.597286$ & $-8.315387$ & $8.707342$ & $-16.048676$ & $-17.384832$ & $57.168652$ & $-25.869081$ & $-57.885948$ \\
\end{tabular}}
\caption{\em \label{table:miniimagenet_energy}\small{Expected energy for different inference methods. Lower energy is better.}}
\end{center}
\end{table}

\begin{table}[h]
\centering
\begin{tabular}{c||c || c}
Method & COCO & ImageNet \\
\hlineB{3}
TRWS & $-28.485636$ & $-28.630786$ \\
AStar & $-28.487422$ & $-28.631678$ \\
Greedy & $-27.246355$ & $-25.496649$ \\
\end{tabular}
\caption{\em \label{table:coco_energy}\small{Mean energy on COCO and ImageNet with $8$ positive and $8$ negative images. Lower energy is better.}}
\end{table}
 
\section{Sharing Parameters of Unary and Pairwise Relation Modules}
As it is discussed in section~\ref{sec:potentials}, both unary and pairwise potential functions use the relation module with an identical architecture. However, since the input class distribution is different for these functions, we choose not to share their parameters. We conduct an experiment to see the effect of parameter sharing in few-shot common object recognition task with $B=5$, $N=8$, and $\bar B = 10$. As Table~\ref{table:miniimagenet} shows, the success rate for this setting is $72.49 \pm 0.98\%$ without parameter sharing. However, when the unary and pairwise are trained with shared relation module parameters, the performance degrades to $69.35 \pm 1.01\%$.
 
\section{More Qualitative Results}
Qualitative results on ImageNet dataset are illustrated in Figure~\ref{fig:imagenet_qualitative}. Figure~\ref{fig:complete_coco_qualitative} shows the complete qualitative results presented in the paper with the negative images on COCO dataset.
\begin{figure*}[ht]
\begin{center}
\includegraphics[width=0.87\textwidth]{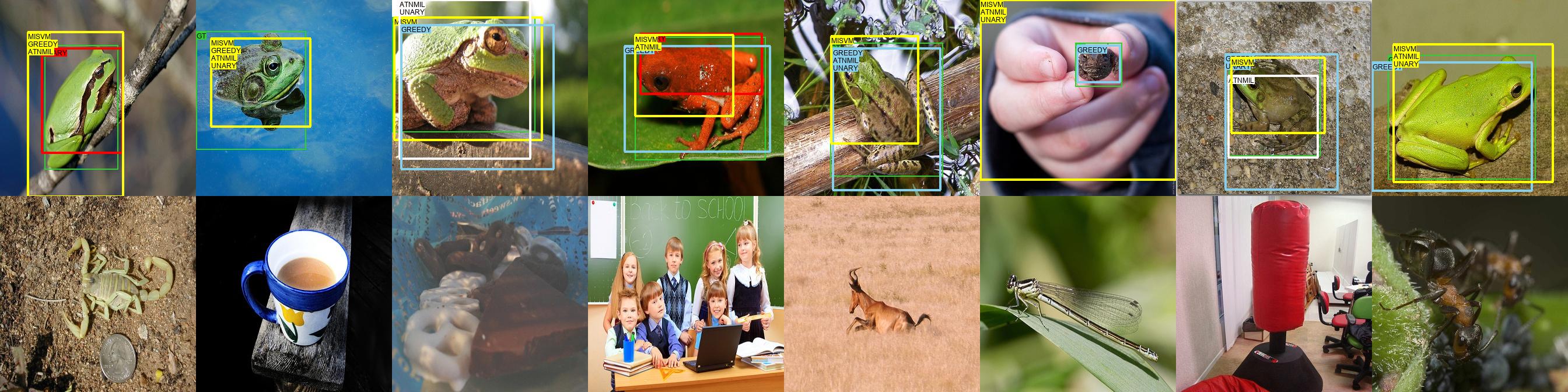} \\\vspace{0.5cm}
\includegraphics[width=0.87\textwidth]{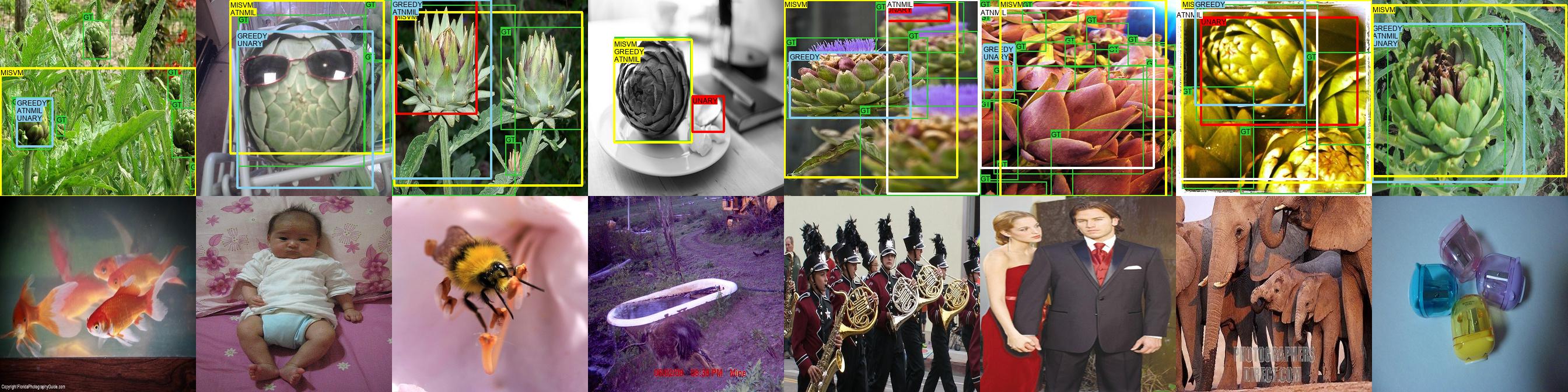} \\\vspace{0.5cm}
\includegraphics[width=0.87\textwidth]{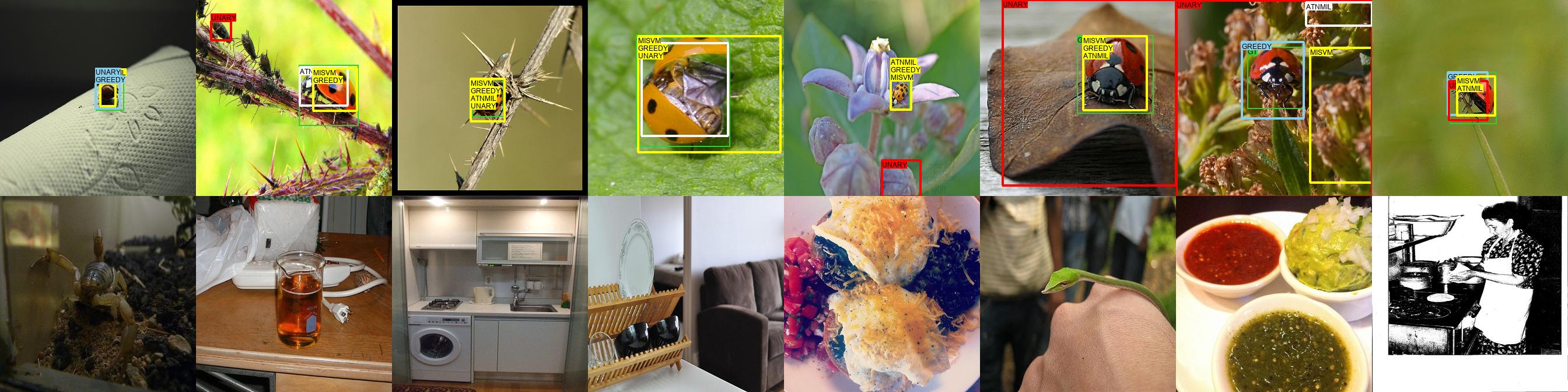} \\\vspace{0.5cm}
\includegraphics[width=0.87\textwidth]{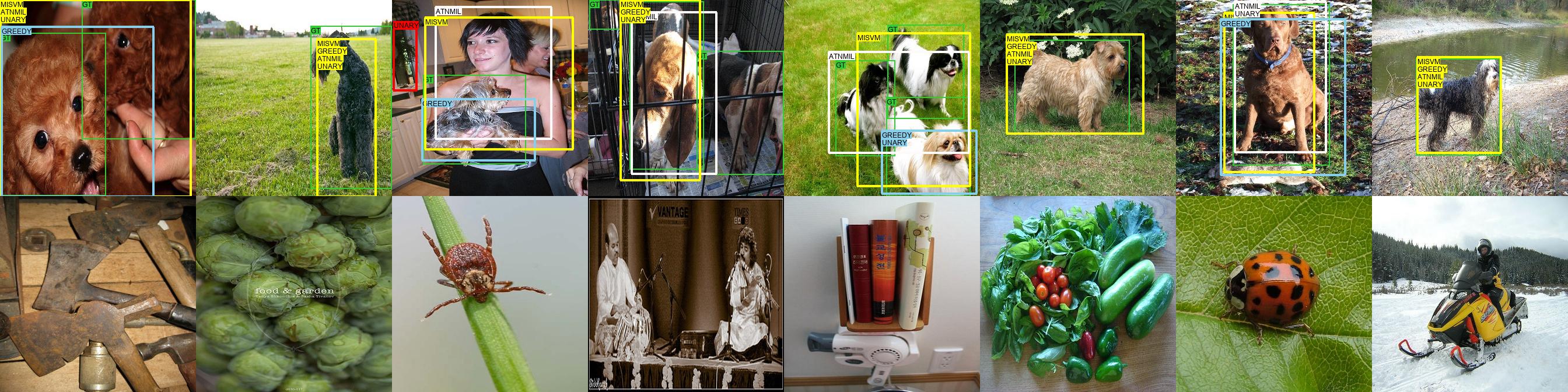} \\\vspace{0.5cm}
\includegraphics[width=0.87\textwidth]{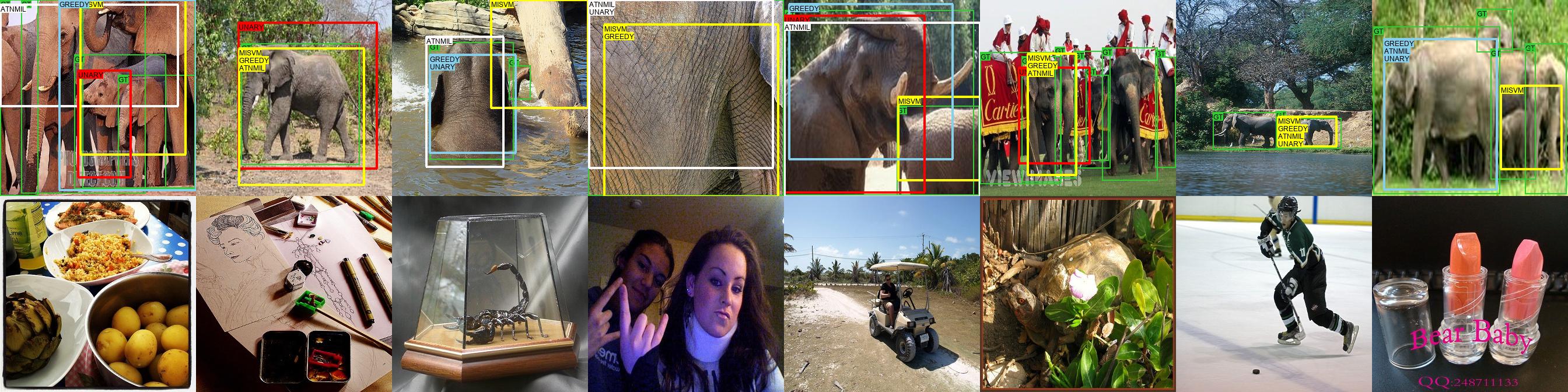}
\caption{\label{fig:imagenet_qualitative}\em Qualitative results on ImageNet dataset. In each problem, the first row and the second row show positive and negative images respectively. While different methods work as good in easier images with one object, the greedy method performs better in harder examples with multiple objects in each image. Selected regions are tagged with method names. Ground-truth target bounding box is shown in green with tag ``GT''.}
\end{center}
\vspace{-0.9cm}
\end{figure*}

\begin{figure*}[ht]
\begin{center}
\includegraphics[width=0.87\textwidth]{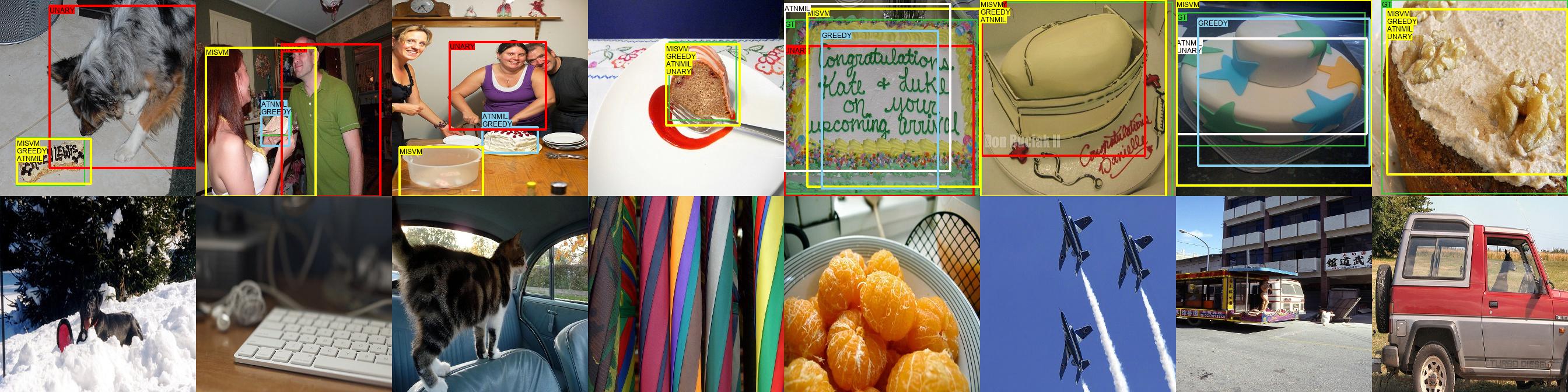} \\\vspace{0.5cm}
\includegraphics[width=0.87\textwidth]{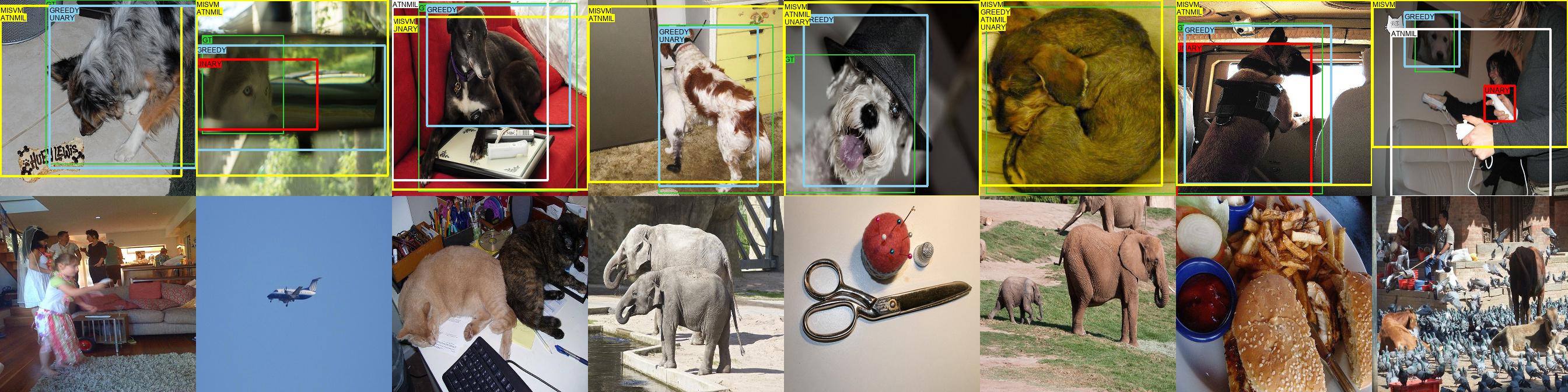} \\\vspace{0.5cm}
\includegraphics[width=0.87\textwidth]{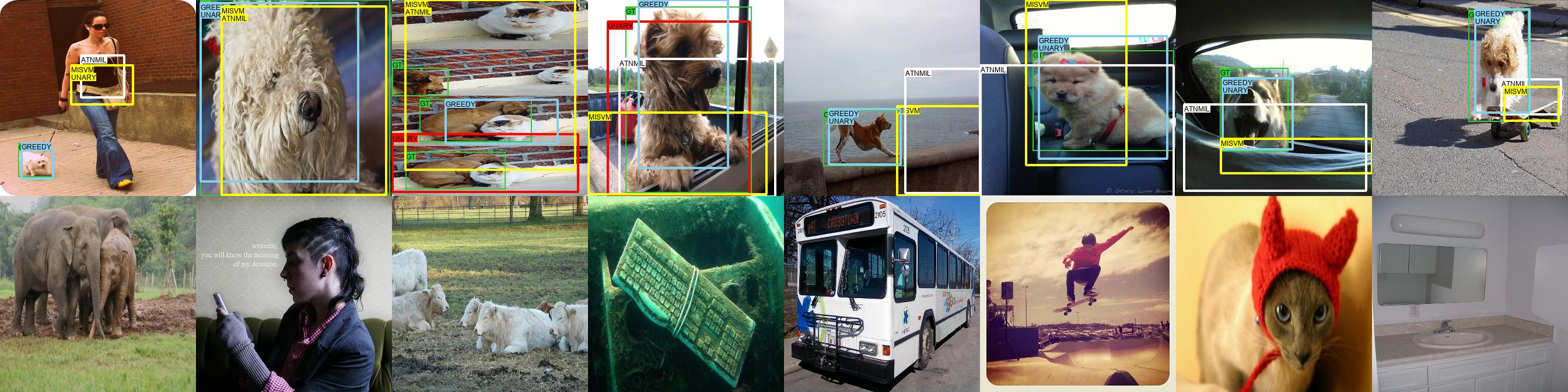} \\\vspace{0.5cm}
\includegraphics[width=0.87\textwidth]{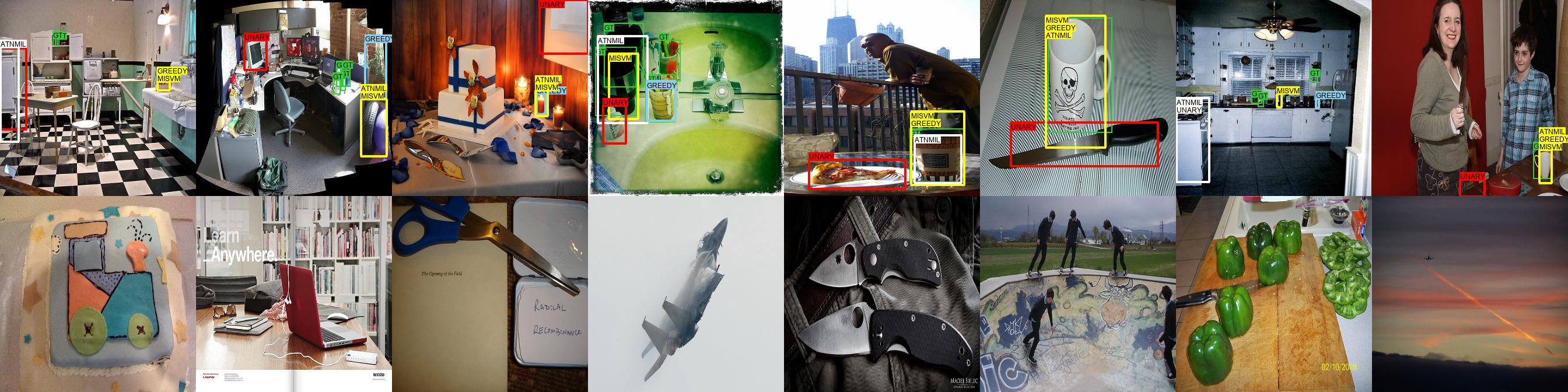}
\caption{\label{fig:complete_coco_qualitative} \em Qualitative results on COCO. Complete version of the results shown in Figure~\ref{fig:coco_qualitative} of the paper with negative images. In the first problem, class ``Person'' does not appear in the negative images. This could explain why ``Unary Only'' method detects people in the first problem.} 
\end{center}
\vspace{-0.9cm}
\end{figure*}

\bibliographystyleRef{plain}
\bibliographyRef{supp_bib}
\fi

\end{document}